\documentclass[%
 reprint,
superscriptaddress,
 amsmath,amssymb,
 aps,
 floatfix
]{revtex4-1}

\usepackage{graphicx}
\usepackage{dcolumn}
\usepackage{bm}

\begin{document}

\preprint{APS/123-QED}

\title{Use of recurrent infomax to improve the memory capability of input-driven recurrent neural networks}

\author{Hisashi \surname{Iwade}}
\email{iwade@acs.i.kyoto-u.ac.jp}
\affiliation{Graduate School of Informatics, Kyoto University, Yoshida Honmachi, Sakyo-ku, Kyoto 606-8501, Japan}
\author{Kohei \surname{Nakajima}}%
\affiliation{Graduate School of Information Science and Technology, University of Tokyo, Tokyo 113-8656, Japan}
\affiliation{JST, PRESTO, 4-1-8 Honcho, Kawaguchi, Saitama 332-0012, Japan}
\author{Takuma \surname{Tanaka}}
\affiliation {Faculty of Data Science, Shiga University, 1-1-1 Banba, Hikone, Shiga 522-8522, Japan}
\author{Toshio \surname{Aoyagi}}
\affiliation{Graduate School of Informatics, Kyoto University, Yoshida Honmachi, Sakyo-ku, Kyoto 606-8501, Japan}

\date{\today}

\begin{abstract}
The inherent transient dynamics of recurrent neural networks (RNNs) have been exploited as a computational resource in input-driven RNNs.
However, the information processing capability varies from RNN to RNN, depending on their properties. Many authors have investigated the dynamics of RNNs and their relevance to the information processing capability.
In this study, we present a detailed analysis of the information processing capability of an RNN optimized by recurrent infomax (RI), which is an unsupervised learning scheme that maximizes the mutual information of RNNs by adjusting the connection strengths of the network.
Thus, we observe that a delay-line structure emerges from the RI and the network optimized by the RI possesses superior short-term memory, which is the ability to store the temporal information of the input stream in its transient dynamics.

\begin{description}


\item[PACS numbers]
89.75.Fb, 84.35.+i, 87.19.lv
\end{description}
\end{abstract}

\pacs{Valid PACS appear here}
\maketitle


\section{Introduction}
\begin{figure*}[ht]
\centering
\includegraphics[width=17.cm,clip]{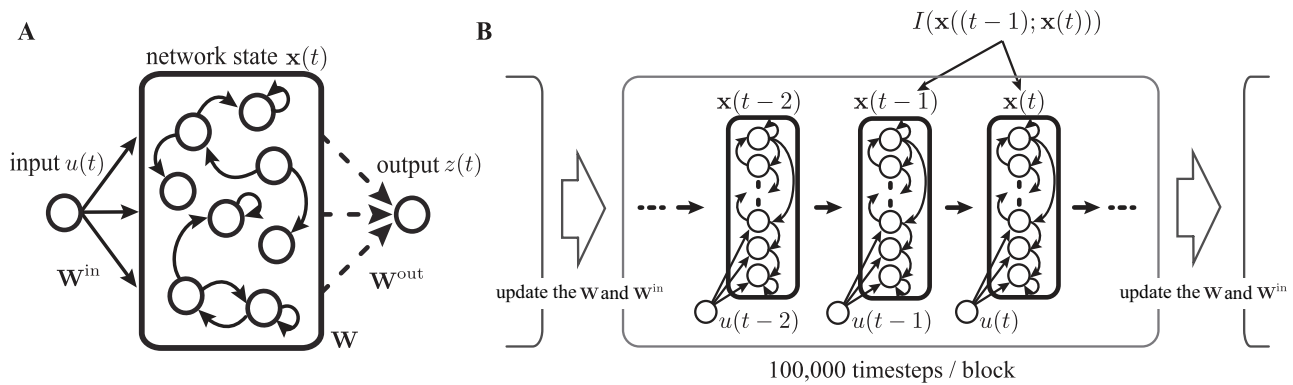}
\caption{(A) The architecture of a conventional RC. 
The connections represend by the solid lines and dashed lines are optimized using RI and the least square method, respectively, for a specific task.
(B) The scheme for RI. The connection weights are updated between two consecutive blocks and each block consists of $100{,}000$ timesteps. The first $50{,}000$ timesteps are for $h_i$ to converge to a steady state and the last $50{,}000$ timesteps are used to calculate mutual information.}
\label{Fig:reservoir}
\end{figure*}

A framework called Reservoir Computing (RC) was proposed as a novel brain-inspired information processing \cite{Maass2002, Jaeger2004a}.
One of interesting features of RC is that it exploits the inherent transient dynamics of recurrent neural networks (RNNs) as a computational resource in input-driven RNNs.
Because of this feature RC has been focused as a theoretical framework to explain the mechanism of information processing of the central nervous system (CNS) \cite{Maass2002, Buonomano2009, rabinovich2008transient}, and RC has been used to emulate the human activity like motor action \cite{Sussillo2009, Laje2013}.
As illustrated in Fig.~\ref{Fig:reservoir}(A), the architecture of RC generally consists of three layers, an input layer, a reservoir layer implemented with a recurrent neural network (RNN), and an output layer.
One of the significant features of this framework is the ability to embed the information of the previous input into the transient dynamics of RNNs, which enables real-time information processing of input streams.
This feature is often called a short-term memory property.
Further, nonlinearity is another feature of the framework that is defined by the ability to emulate nonlinear information processing.
Additionally, the supervised learning of the reservoir layer is not required if the dynamics of reservoir layer is appropriate to required dynamical systems.
The connection weights from the RNNs to the output, which are called readout weights, are adjusted to learn the required dynamical systems.
Although RC has a simple learning scheme, it has been observed to exhibit high performance in many machine learning tasks \cite{Antonelo2008, Jalalvand2015, Salmen2005, Skowronski2007, Jaeger2002}.

The information processing capability of an RC system is dependent on the inherent dynamics of the reservoir layer.
Consequently, many authors have investigated the information processing capabilities of various dynamical systems.
One of the most well known networks is a dynamical system at the edge of chaos (or at the edge of stability), which is the transition point from ordered to chaotic dynamics.
Such systems exhibit superior information processing capability \cite{Bertschinger2004a, Toyoizumi2011}.
Some studies have investigated the information processing capabilities of physical phenomena.
These approaches exploit the dynamics of physical phenomena as reservoirs, such as the surface of water \cite{fernando2003pattern}, optoelectronic and photonic systems \cite{Larger2012, appeltant2011information, woods2012optical}, ensemble quantum systems \cite{PhysRevApplied.8.024030}, neuromorphic chips \cite{stieg2012emergent}, and the mechanical bodies of soft robots \cite{nakajima2013soft, nakajima2014exploiting, Nakajima2015}.

In this study, we exploit the RNNs that are optimized by an optimization principle called recurrent infomax (RI)  \cite{Tanaka2008} and investigate the information processing capabilities of these networks.
RI was proposed as an extension of feedforward infomax, which was originally proposed by Linsker \cite{Linsker1988a}.
RI provides an unsupervised learning scheme to maximize information retention in an RNN, which is quantified with mutual information.
The study which introduced RI explained that the RNNs optimized by RI exhibit the characteristic dynamics of neural activities such as cell assembly-like and synfire chain-like spontaneous activities as well as critical neuronal avalanches in CNS \cite{Tanaka2008}.
An RNN optimized by RI is expected to exhibit superior information processing capability because the principle of RI is to maximize the information retention.
However, the method by which RI affects the information processing capability of an RNN remains unclear.
In this study, we investigate to clarify this method in detail based on the framework of RC.
Initially, we optimize RNNs using RI and analyze the basis of their information processing capabilities using benchmark tasks.
\clearpage

\section {Model}
We consider an input-driven network consisting of $N$ neurons where the state of each neuron $x_i(t) \in \{0,1\}\ (i = 1,... ,N)$ is updated synchronously and stochastically at discrete timesteps.
The firing probability of neuron $i$ is determined by its interaction with another neuron $j \ (j = 1,... ,N)$ that is connected with the weight $W_{ij}$ 
and the input $u(t) \in \{0,1\}$ that is connected with the weight $W_{i}^{\mathrm{in}}$ as
\begin{align}
&p(x_i(t+1)=1) =\frac{p_{\mathrm{max}}}{1 + \exp(-U_i(t))}, \\ 
&U_i(t) = \sum^N_{j=1}W_{ij}(x_j(t) -\bar{p}_j) + W_i^{\mathrm{in}}(u(t)-\bar{p}_0) - h_i(t),
\end {align}
where $p_{\mathrm{max}}$ is the maximal firing probability, $\bar{p}_0$ is the expected input value, $\bar{p}_j$ is the mean firing rate of neuron $j$ and $h_i(t)$ is the bias for neuron $i$.
The firing frequency of each neuron is indicated by the mean firing rate.
For example, the neuron $i$ is activated once in $10$ timesteps on an average at $\bar{p}_i=0.1$.
The maximal firing probability $p_{\mathrm{max}}$ indicates the reliability of the activation in response to the input.
In this study, the maximal firing probability and the mean firing rate are fixed at $p_{\mathrm{max}} = 0.8 $ and $\bar{p}_i = 0.1$, respectively.
The input $u(t)$ is sampled randomly from $\{0, 1\}$ with the expected value $\bar{p}_0=0.5$. 
The bias $h_i(t)$ is used to fix the average firing probability $\bar{p_i}$ and is updated at each timestep by
\begin {align}
h_i(t+1) & = h_i(t) + \epsilon ( x_i(t+1) - \bar{p}_i),
\end {align}
where $\epsilon$ is the learning rate, which remains constant at 0.01. 
The initial conditions of the network ($W_{ij}$ and $W_i^{\mathrm{in}}$) are sampled from the Gaussian distribution with $\mu = 0$ and $\sigma^2 = 0.01$. 

\section {Recurrent Infomax}
Further, we briefly describe the algorithm of RI \cite{Tanaka2008}.
In this study, RI is applied to the connection weights within the RNN and from the input to the RNN.
The connection weights are represented by solid lines in Fig.~\ref{Fig:reservoir}(A).
As illustrated in Fig.~\ref{Fig:reservoir} (B), the connection weights are updated at the end of each block, which consists of $100{,}000$ timesteps.
The connection weights at the $b+1^{\mathrm{th}}$ block $\textbf{W}(b+1)$ are calculated using the steepest gradient method as follows:
\begin{equation}
W_{ij}(b+1)  = W_{ij}(b) + \eta \frac{\partial I(b)}{\partial W_{ij}},
\end{equation}
where $\eta$ is the learning rate that remains constant at $\eta = 0.2$ throughout this study.
The input connection weights $W_i^{\mathrm{in}}$ and the input $u(t)$ are represented by $W_{i0}$ and $x_0(t)$, respectively.
The gradient of the mutual information with respect to $W_{ij}$ is formulated using several approximations \cite{Tanaka2008}.
The mutual information between two successive states at the $b^{\mathrm{th}}$ block is denoted using Gaussian approximation and is represented as follows:
\begin{equation}
I(b) = \log|{\textbf{C}}| - \frac{1}{2}\log|\textbf{D}|,
\end{equation}
where
\begin{gather}
\textbf{C} = \left(
\begin{array}{ccc}
E_{00} & \ldots & E_{0N} \\
\vdots & \ddots & \vdots \\
E_{N0} & \ldots & E_{NN}
\end{array}
\right),  \\[1em]
\textbf{D} = \left(
\begin{array}{cccccc}
E_{00} & \ldots & E_{0N} & E_{0\widehat{0}} & \ldots & E_{0\widehat{N}} \\
\vdots & \ddots & \vdots & \vdots & \ddots & \vdots \\
E_{00} & \ldots & E_{0N} & E_{N\widehat{0}} & \ldots & E_{N\widehat{N}} \\
E_{\widehat{0}0} & \ldots & E_{\widehat{0}N} & E_{\widehat{0}\widehat{0}} & \ldots & E_{\widehat{0}\widehat{N}} \\
\vdots & \ddots & \vdots & \vdots & \ddots & \vdots \\
E_{\widehat{N}0} & \ldots & E_{\widehat{N}N} & E_{\widehat{N}\widehat{0}} & \ldots & E_{\widehat{N}\widehat{N}}
\end{array}
\right),
\end{gather}
\begin{subequations}
\begin{align}
E_{ij} & =  \frac{1}{T}\sum(x_i(t) - \bar{p}_i)(x_j(t) - \bar{p}_j), \\
E_{\widehat{i}j} & =  \frac{1}{T} \sum(x_i(t + 1) - \bar{p}_i)(x_j(t) - \bar{p}_j), \\
E_{i\widehat{j}} & = \frac{1}{T} \sum(x_i(t) - \bar{p}_i)(x_j(t + 1) - \bar{p}_j), \\
E_{\widehat{i}\widehat{j}} & = \frac{1}{T} \sum(x_i(t + 1) - \bar{p}_i)(x_j(t + 1) - \bar{p}_j).
\end{align}
\end{subequations}

\begin{figure*}[th]
\centering
\includegraphics[width=17cm]{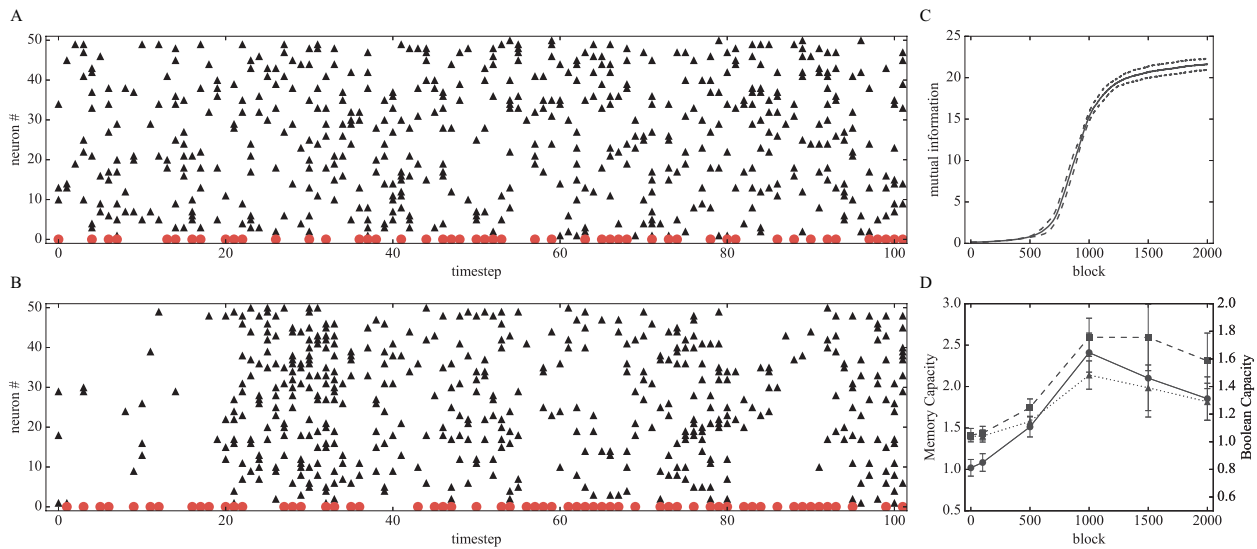}
\caption{(Color online)
(A, B) The raster plots for the firing activity of the network at the $0^{\mathrm{th}}$ and the $1000^{\mathrm{th}}$ blocks from $50{,}000$ to $50{,}100$ timesteps.
The circular and triangular points indicate the firing of the input and the neurons in RNN.
(C) The result of mutual information as a function of the block.
The solid line indicates the mean for the different initial connection weights, whereas the dotted line indicates the standard deviation for the networks with the different initial conditions.
The number of trials is $10$, which remains constant throughout this study 
(D) The results of MC and BCs as a function of the block.
The solid line indicates the result of the memory capacity task.
The dashed and dotted lines indicate the result of a $2$-bit and $3$-bit Boolean function task, respectively.
The error bar indicates the standard deviation for networks with different initial conditions.}
\label{Fig:single_res}
\end{figure*}

Finally, the gradient of the mutual information with respect to the connection weights is approximated by
\begin{widetext}
\begin{align}
\frac{\partial I(b)}{\partial W_{ij}} \approx &  \frac{1}{2} \sum(1 - \delta_{ij})(E_{\widehat{i}\widehat{k}}E_{\widehat{j}l} + E_{\widehat{i}l}E_{\widehat{j}\widehat{k}})(2(C^{-1})_{ji} - (D^{-1})_{ji} - (D^{-1})_{j+Ni+N}) \nonumber \\
& - \frac{1}{2}((1-2\bar{p}_k)(1-2\bar{p}_l)E_{\widehat{k}l} + \bar{p}_k\bar{p}_l(1 - \bar{p}_k)(1 - \bar{p}_l))((D^{-1})_{lk+N})+(D^{-1})_{k+Nl}).
\end{align}
\end{widetext}
Typical instances of the firing activity of the network optimized by RI at the $0^{\mathrm{th}}$ and $1500^{\mathrm{th}}$ block are depicted in Figs. \ref{Fig:single_res}(A) and \ref{Fig:single_res}(B).
The neurons in the network optimized by RI tend to fire simultaneously.
Similar dynamics were also exhibited in \cite{Tanaka2008}.
As depicted in Fig.~\ref{Fig:single_res}(C), the mutual information is increased by RI and it is almost saturated at the $1500^{\mathrm{th}}$ block.
Therefore, we terminate RI at the $1500^{\mathrm{th}}$ block for our study.

\section{Information processing capability}
We use three benchmark tasks, memory capacity, $2$-bit Boolean function, and $3$-bit Boolean function tasks to evaluate the information processing capability of a network optimized using RI.
Each benchmark task consists of washout ($50{,}000$ timesteps), learning ($1500$ timesteps), and testing ($1500$ timesteps) phases.
The washout phase is used to eliminate the influence of the initial state of the network and to converge the bias $h_i(t)$ to a steady-state value.
Further, the learning phase is used to learn readout weights $\textbf{W}_\tau^{\mathrm{out}}$, whereas the testing phase ($1500$ timesteps) is used to evaluate the performance of each task accordingly.

The memory capacity task has been extensively used as a benchmark task in the framework of RC \cite{Bertschinger2004a, Jaeger2002a}.
This task evaluates the extent of decoding the information of past inputs from the network state with memoryless readout.
The performance of this task represented by $\mathrm{MC}$ is the summation of $\tau$-delay memory functions [$\mathrm{MF}_\tau$ $(\tau = 1, ..., 50)$] defined by
\begin {equation}
{\mathrm{MC}}   =  \sum^{50}_{\tau=1}{\mathrm{MF}}_\tau.
\end{equation}
The memory function $\mathrm{MF}_\tau$ is the determination coefficient between $u(t-\tau)$ and $z_\tau(t) = \textbf{W}_\tau^{\mathrm{out}}\textbf{x}(t)$ defined by
\begin {equation}
{\mathrm{MF}}_\tau =  \max_{W_\tau^{\mathrm{out}}}\frac{{\mathrm{cov}}^2(z_\tau(t),u(t-\tau))}{\sigma^2(z_\tau(t))\sigma^2(u(t-\tau))}.
\end {equation}
The readout weights, $\textbf{W}_\tau^{\mathrm{out}}$, are trained in accordance with $\textbf{W}_\tau^{\mathrm{out}} = (\textbf{X}^{\text{T}}\textbf{X})^{-1}\textbf{X}^{\text{T}}\textbf{y}$,
where $\textbf{y}$ is the vector of $\tau$-timestep past inputs and $\textbf{X}$ is the matrix of network states 
during the learning phase.

The $n$-bit ($n=2$, $3$) Boolean function task evaluates whether the network can process the Boolean operation using previous $n$-bit inputs.
We have depicted an example in Table.~1.
Each rule of the Boolean function tasks requires separate information processing.
Hence, the execution of each rule may provide us with useful insights about the property of information processing for RNNs optimized by RI. 
The total number of rules is $2^{n^2}$, and all rules except for two, are applied.
Exceptional rules are those for which output $f_{\tau}$ is $0$ or $1$ regardless of the input.
Thus, the number of rules, $K$, for each $n$-bit Boolean function task is $2^{n^2}-2$.

\begin{table}[h]
\begin{center}
\caption{An example of a rule table for a 2-bit Boolean function.}
\begin{tabular}{|c|c||c|} \hline
$u(t -\tau)$ & $u(t-\tau-1)$ & $f_{\tau}$(t) \\ \hline \hline
0 & 0 & 1 \\
0 & 1 & 0 \\
1 & 0 & 1 \\
1 & 1 & 0 \\ \hline
\end{tabular}
\end{center}
\end{table}

The performance of this task is termed Boolean capacity ($\mathrm{BC}$), which is the summation of Boolean function $\mathrm{BF^k_{\tau}}$ at each rule $k$ $(k = 1, ..., K)$ and time delay $\tau$ $(\tau = 1, ..., 50)$ and is represented as follows:
\begin {align}
{\mathrm{BC}} & =  \frac{1}{K}\sum^{K}_{k=1}\sum^{50}_{\tau=1}{\mathrm{BF}}_{\tau}^k,
\end {align}
where
\begin {align}
{\mathrm{BF}}_{\tau}^k & =  \max_{W_{\tau}^{k, \mathrm{out}}}\frac{{\mathrm{cov}}^2(z_k(n),f_{\tau}^k(n))}{\sigma^2(z_k(n))\sigma^2(f_{\tau}^k(n))}.
\end {align}

The results of the benchmark tasks are depicted in Fig.~\ref{Fig:single_res}(D).
We suspended the update and performed the benchmark tasks for the networks while evaluating the information processing capability of the networks during RI.
The process of updating RI was resumed, after evaluating the performance of benchmark tasks.
Though both the memory capacity and Boolean capacities are improved using RI, they exhibit a peak at the $1000^{\mathrm{th}}$ block.
This is in sharp contrast to the mutual information, which is observed to saturate at about the $1500^{\mathrm{th}}$ block.

\begin{figure}[h]
\centering
\includegraphics[width=8cm]{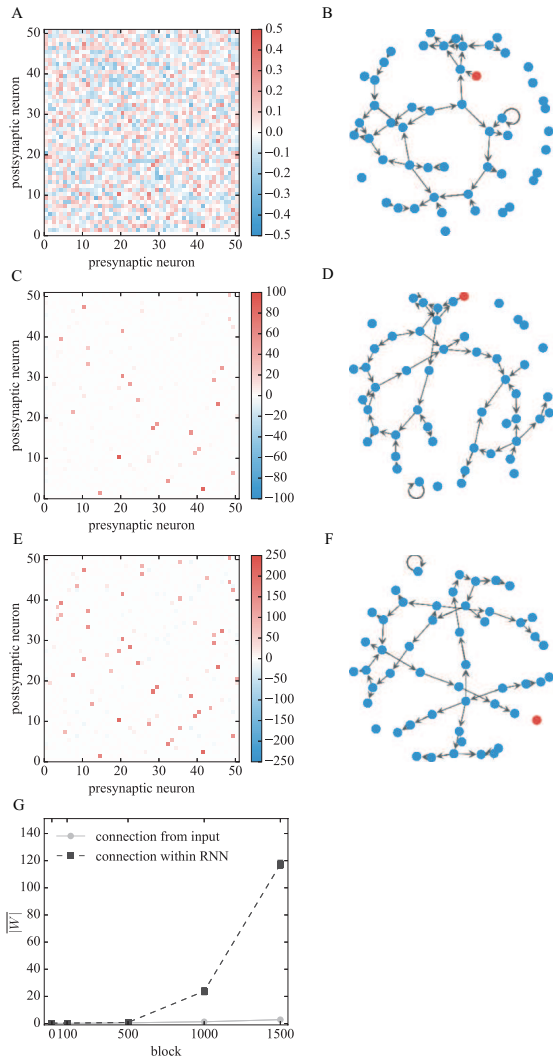}
\caption{(Color online)
(A, C, E) The heatmap for the connection weights at the $0^{\mathrm{th}}$, $1000^{\mathrm{th}}$ and $1500^{\mathrm{th}}$ blocks.
The neuron ${\mathrm{\#}}0$ indicates the input.
(B, D, F) The network structure is designed using each left heatmap.
The connections with the 50 largest absolute values are visualized. The blue and red nodes represent the neurons in the RNN and the input, respectively.
(G) The mean of the absolute strength of the connection weights from the input to the RNN and those within the RNN as a function of the block.
The mean and standard deviations of the networks under distinct initial conditions are illustrated, but the error bar is not substantial due to the small standard deviation.}
\label{network}
\end{figure}

We visualized the network structure to investigate the reason that both $\mathrm{MC}$ and $\mathrm{BC}$s exhibit a peak.
The network structure is depicted in Fig.~\ref{network}(A-F), which represents the connections with the $50$ largest absolute values.
Most of the visualized connections are those within the RNN, and a few are connections from the input to the RNN.
Additionally, we calculated the mean of the absolute connection weights ($\overline{W}$) within the RNN and those from the input to the RNN, as illustrated in Fig.~\ref{network}(G).
The former is calculated using the $50$ largest absolute values in $\textbf{W}$ while the latter is calculated using $\textbf{W}^{\mathrm{in}}$.
We can confirm that the value of  $\overline{W}$ within the RNN become much higher than that of $\overline{W}$ from input to the RNN as the block of RI increases.
This result suggests that the information of the input is not stored in the network, but the only information existing in the network is preserved when the network is optimized by RI for long term.
Consequently, the performances of both $\mathrm{MC}$ and $\mathrm{BC}$s become worse.

\section{Introduction of input multiplicity}
\subsection{Recurrent Infomax and Information Processing Capability}
In the previous section, we depicted that the information processing capability exhibits an increase at the beginning of RI but peaked at the $1000^{\mathrm{th}}$ block because the connection weights within the RNN became stronger than the input connection weights.
Hence, the input information may not be preserved in the network.
We tried to increase the number of input neurons having common input information to handle this problem.
This is equivalent to introducing the weight coefficient ($K$), which is entitled input multiplicity, in the updated formula of the connection weights from the input to the RNN.
The updated formula is
\begin{equation}
W_{i0}(b+1) = W_{i0}(b) + \eta K \frac{\partial I(b)}{\partial W_{i0}},
\end{equation}
where $ K = \{K|K \in \mathbb{Z},2 \leq K \leq 35\}$.
We optimize the input-driven RNN by RI using the above formula to update the connection weights from the input to RNN and subsequently executed the memory capacity and Boolean function tasks. 
The parameter values except for the weight coefficient $K$ and benchmark tasks of RI are similar to those mentioned in previous section.

\begin{figure}[h]
\centering
\includegraphics[width=8.5cm]{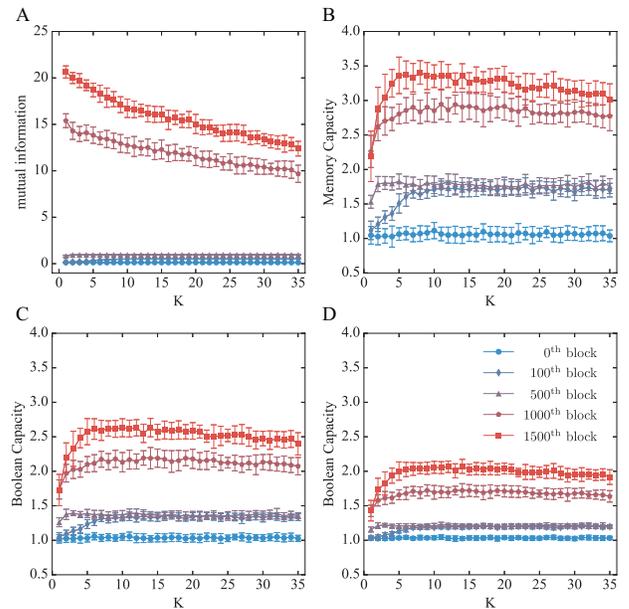}
\caption{(Color online)
(A) The mutual information as a function of input multiplicity, $K$. The color indicates the block number.
(B) The performance of the memory capability task.
(C) The performance of the 2-bit Boolean function task.
(D) The performance of the 3-bit Boolean function task.
}
\label{Fig:multi_k}
\end{figure}

Initially, we provide an overview of the results for the mutual information and benchmark tasks.
As depicted in Fig.~\ref{Fig:multi_k}(A), a large $K$ value decreases the mutual information at the learning phase, although RI increases the mutual information for all values of $K$, a large $K$ decreases the mutual information at the end of the learning phase.
Furthermore, the memory capacity illustrated in Fig.~\ref{Fig:multi_k}(B) also increases throughout the optimization using RI at all values of $K$, except for $K=1$, is maximized around $K=7$.
The results of the $2$-bit and $3$-bit Boolean function tasks illustrated in Fig.~\ref{Fig:multi_k}(C) exhibit similar tendencies as that of the memory capacity task.

\begin{figure}[t]
\centering
\includegraphics[width=8.5cm]{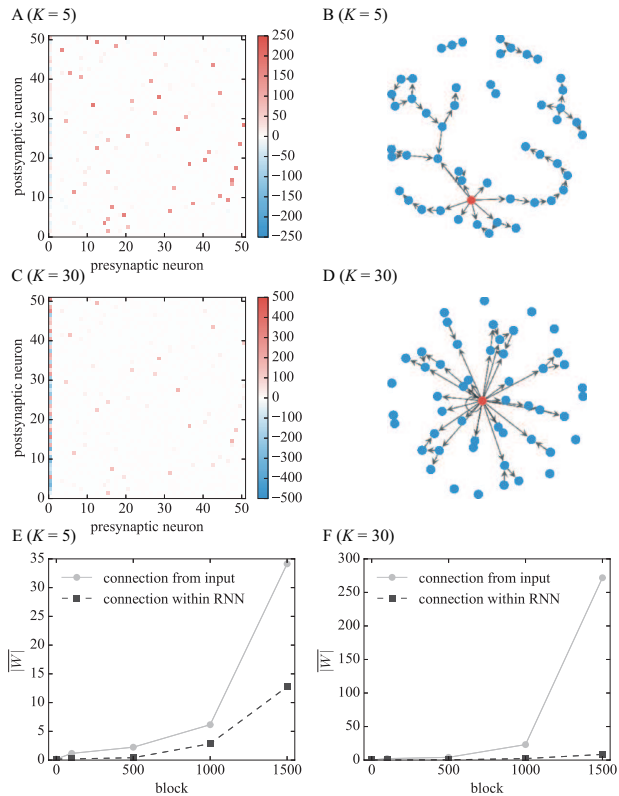}
\caption{(Color online)
(A, C) The heatmap for the connection weights for $K=5$ and $30$ at the $1500^{\mathrm{th}}$ block. Neuron ${\rm \#}0$ indicates the input.
(B, D) The network structure is designed using each left heatmap. The 50 largest absolute connection weights are visualized. The blue nodes represent the neurons in the RNN, whereas the red nodes represent the input.
(E, F) The mean of the absolute strength of the connection weights from the input to the RNN and those within the RNN is a function of the block number.}
\label{Fig:topology}
\end{figure}

Figures.~\ref{Fig:topology}(A-D) illustrate typical network structures at $K=5{,}~30$.
At $K=5$, the strong connection weights from the input to the RNN and those within the RNN coexist.
The postsynaptic input neurons become the presynaptic neurons of other neurons.
Hence, a delay-line structure originates from the input.
Conversely, the strong connection weights are mostly observed to be the connections from the input to the RNN at $K=30$.
The mean strength of the connection weights as a function of the block exhibits opposite tendency to that of connection weights mentioned in the previous section, which increases more rapidly than those within the RNN, as depicted in Fig.~\ref{Fig:topology}(E, F).
These network structures suggest that for a large $K$ value, the input information at the final timestep may be stored dominantly in the network and that majority of the previous input may be lost.

\begin{figure}[h]
\centering
\includegraphics[width=8.5cm]{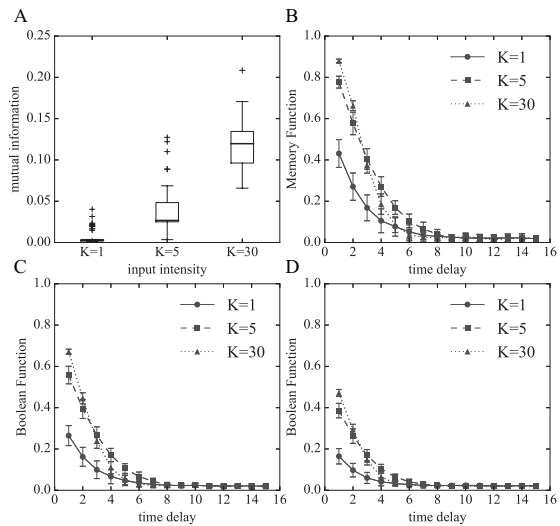}
\caption{
(A) The boxplot of mutual information between last input and the RNN. It is estimated by sampling the $50,000$ timesteps after washout timesteps ($50,000$ timesteps). 
(B) The performance of the memory capacity task as a function of time delay for $K=1$, $5$ and $30$.
(C) The performance of the 2-bit Boolean function task as a function of time delay for $K=1$, $5$ and $30$. 
(D) The performance of the 3-bit Boolean function task as a function of time delay for $K=1$, $5$ and $30$. 
The error bar indicates the standard deviation for the networks under the different initial conditions.}
\label{Fig:mitd}
\end{figure}

We confirm this speculation quantitatively by investigating the mutual information between the input at the last timestep and each neuron in RNN, $I(x_i(t);u(t-1))$ $(i = 1,... ,N)$ as plotted in Fig.~\ref{Fig:mitd}(A) using boxplot.
This result indicates that the network optimized by RI at large values of $K$ contains information about the input given at the last timestep.
The  results of the memory, $2$-bit Boolean, and the $3$-bit Boolean functions as a function of time delay are plotted in Fig.~\ref{Fig:mitd}(B, C).
The performance of the network when $K=30$ at $\tau \leq 2$ is observed to be higher than that when $K=5$.
As the time delay $\tau$ becomes longer, the performance of the network when $K=30$ is inferior to the performance when $K=5$.
All theses results are consistent with the speculation that the network optimized by RI with a large $K$ value can only store the recent input information.
As discussed in the preceding sections the learning with RI using a small $K$ value optimizes the network such that the last input information is not stored dominantly, but the information about its past state is stored.\par

\subsection {Comparison Of Information Processing Capability Between Linear And Nonlinear Rules}
\begin{figure}[t]
\centering
\includegraphics[width=8.5cm]{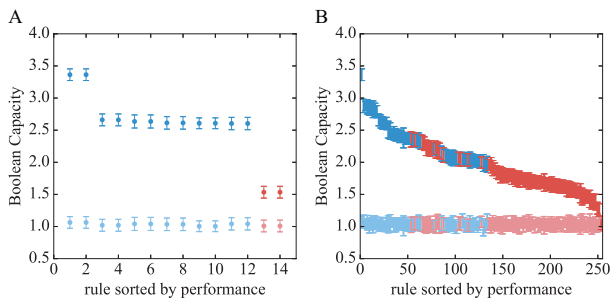}
\caption{(Color online)
The performance of each rule of the Boolean function tasks for $K=5$ at the $0^{\mathrm{th}}$ block (the points with light colors) and $1500^{\mathrm{th}}$ block (the points with deep colors). The blue points indicate the linear rule while the red indicate the nonlinear rule.  The error bar indicates the standard deviation for the different initial networks. (A) $2$-bit and (B) $3$-bit Boolean function tasks.
}
\label{Fig:Boolean}
\end{figure}

We classified the rules of the Boolean function tasks into two classes, i.e., linear and nonlinear rules, to investigate the property of information processing of the network optimized by RI.
The classification criterion is based on whether the rule is linearly separable or not \cite{Chua2002}.
Figure \ref{Fig:Boolean} plots the performance of each rule sorted by score and colored to denote whether it is linear or nonlinear.
Initially, we discuss the results of the $2$-bit Boolean function task depicted in Fig.~\ref{Fig:Boolean}(A).
The performance of the majority of the rules are improved using RI, and the networks exhibit high performance in the execution of linear rules as compared to nonlinear rules.
This result suggests that the network optimized by RI performed the linear information processing better than nonlinear information processing. 
The initial two rules were performed more efficiently than the other linear rules because the difficulty of executing a rule is dependent on both linear separability as well as short-term memory.
These two rules are identical to the memory capacity task, and which does not require the information of $u(t-\tau-1)$ but it only requires the information of $u(t-\tau)$.
Therefore, these rules are comparatively simple to execute than the other linear rules.

The results of the $3$-bit Boolean function task illustrated in Fig.~\ref{Fig:Boolean}(B) exhibit the tendency that linear rules are performed better than nonlinear rules.
However, some nonlinear rules are performed better than linear rules. Consequently, the difference between the two kinds of rules, which is observed in the $2$-bit Boolean function task, is unclear, because some rules require previous information about the input $u(t-\tau-2)$ while the performance of each rule may be influenced not only by nonlinearity but also the short-term memory in the $3$-bit Boolean function task.

\section{Discussions}
In this study, we optimized the input-driven RNN using RI and evaluated the information processing capability of the network using the memory capacity and the Boolean function tasks.
Though naive RI did not improve the information processing capability, RI with input multiplicity improves the performances of the memory capacity and Boolean function tasks because the connection weights of the input increase preferentially and the input information is stored in the network.
We tackled this problem by introducing input multiplicity.
An appropriate input multiplicity optimizes the network to store the information about the past input, and the network partially acquires a delay-line structure, which is reported to be one of the structure that can achieve optimal short-term memory \cite{Jaeger2002a, Ganguli2008, Rodan2010}.
However, our model is stochastic, and the complete delay-line network may not be suitable for short-term memory because the optimal short-term memory of the delay-line network is based on an assumption that the information of presynaptic neurons is conveyed to postsynaptic neurons without any loss, and the retention of the input information of the delay-line network is fragile to the unreliable firing of neurons.

The performance for each rule of the $2$-bit Boolean function task clearly demonstrate the information processing property of the network optimized by RI. 
Although the performance for majority of the rules is improved by RI, the degrees of improvement depend majorly on the linearity of the rules.
The linear rules are processed in an efficient manner in contrast to the nonlinear rules which include exclusive OR operations, i.e., $u(t-\tau) \ \otimes \ u(t-\tau-1)$.
The tradeoff between short-term memory and nonlinearity can be observed in multiple dynamical systems such as logistic maps, diffusion equations and echo state network \cite{Dambre2012}.
Though there is no theoretical proof about this tradeoff, it is possible that RI may optimize the network under this tradeoff constraint.
Thus, the short-term memory of the network that is optimized by RI may be superior to that of nonlinearity.

We clarify the relationships between the network optimized by RI and its information processing properties under constrained system.
The dynamics of the network to which current RI can be applied is limited to stochastic dynamics under Gaussian approximation.
Therefore, it is interesting to extend RI to other stochastic or deterministic dynamics of the networks and investigate the relationships between RI and information processing properties of RNN.


\section{Acknowledgments}
This work was supported by MEXT KAKENHI Grant Numbers 15H05877 and 26120006; JSPS KAKENHI Grant Numbers 16K16123, 16KT0019, 15587273 and 15KT0015; and JST PRESTO Grant Numbers JPMJPR15E7, Japan, KAKENHI No. 15K16076, No. 266880010 and No. 16777928.


%

\end{document}